\documentclass{article} 
\usepackage{iclr2021_conference,times}

\usepackage{amsmath,amsfonts,bm}

\def\eqref#1{equation~\ref{#1}}

\def\1{\bm{1}}

\DeclareMathAlphabet{\mathsfit}{\encodingdefault}{\sfdefault}{m}{sl}
\SetMathAlphabet{\mathsfit}{bold}{\encodingdefault}{\sfdefault}{bx}{n}

\usepackage{hyperref}
\usepackage{url}
\usepackage{graphicx}
\usepackage{multirow}
\usepackage{geometry}
\usepackage{float}
\usepackage{subfig}

\title{ Fashion Recommendation: Outfit Compatibility using GNN}

\author{Samaksh Gulati
\\
\texttt{\{sgulati\}@gatech.edu} \\
}

\iclrfinalcopy 
\begin{document}

\maketitle

\begin{abstract}

Numerous industries have benefited from the use of machine learning, and the fashion industry is no exception. By gaining a better understanding of what makes a "good" outfit, companies can provide useful product recommendations to their users. In this project, we follow two existing approaches that employ graphs to represent outfits and use modified versions of the Graph neural network (GNN) frameworks. Both Node-wise Graph Neural Network (NGNN)\cite{Cui_2019} and Hypergraph Neural Network (HGNN)\cite{fashion_hypergraph} aim to score a set of items according to the outfit compatibility of items. The data used is the Polyvore Dataset which consists of curated outfits with product images and text descriptions for each product in an outfit.  We recreate the analysis on a subset of this data and compare the two existing models on their performance on two tasks -- (1) Fill-in-the-blank (FITB): finding an item that completes an outfit, and (2) Compatibility prediction: estimating compatibility of different items grouped as an outfit. We are able to replicate the results directionally and find that HGNN does have a slightly better performance on both tasks. On top of replicating the results of the two papers we also tried to use embeddings generated from a vision transformer (ViT) and witness enhanced prediction accuracy across the board.

\end{abstract}

\section{Introduction}




The objective of this paper is to explore the use of different graph-based frameworks for the representation of clothing/accessory items and outfits in the task of fashion recommendation. We aim to tackle the practical problem of fashion recommendation, specifically what item matches and compliments an outfit. The main aim is to estimate an outfit compatibility score, a quantitative metric to measure how well different items in an outfit complement each other. To do this, we leverage two existing implementations of Graph Neural Networks (GNNs) to create embeddings of all items in the same outfit. We then compare the performance of these two techniques against each other. We also look at how different modalities of representing items influence performance.

We use the Polyvore dataset which consists of thousands of curated outfits by experts. Specifically it contains not only the images of the products but also their textual descriptions and other metadata. 


We use two different methods involving GNNs to create image and text embeddings. A GNN learns a target node’s representation by propagating the neighbouring information in an iterative manner thus ensuring similarity of representation of items linked closely.

\begin{itemize}
    \item The first approach called Node-wise Graph Neural Network (NGNN) \cite{Cui_2019} improves on GNN by eschewing parameter sharing to capture item characteristics better. It uses the category of the item as a mainstay in message propagation by using weighted edges based on the category of nodes and also using category-specific item representation. Finally, it uses an attention mechanism to score the outfit.
    
    \item The second approach, HGNN (Hypergraph-GNN) \cite{fashion_hypergraph}, uses hyperedges to denote each outfit. A hypergraph is a generalization of a graph in which an edge can join any number of nodes. Using this approach, theoretically, captures more complex interactions between each item in an outfit leading to key features being captured accurately in the embeddings.

\end{itemize}

We used two tasks to validate our models, namely (1) Fill-In-The-Blank (FITB) tasks where an item is selected from multiple outfit choices and (2) Compatibility Prediction tasks where we predict of the compatibility score of a group of items. 

The experimental results, which are presented in Results section firstly, show that the HGNN representation is marginal improvement over the NGNN in terms of accuracy by about 1\% and 11\% in the fill-in-the-blank and compatibility prediction tasks, respectively. Secondly, the accuracy and AUC metrics are better using a multi-modal approach rather than text-only or visual-only approaches.

The fashion graphs of NGNN and HGNN frameworks demonstrated shows that the potentially costly and tedious work of putting together an outfit and making a fashion decision can be automated to some extent. Utilizing these tools can help companies in the fashion industry provide item recommendations for their customers on online platforms, leading to increased usability, buying desire, and subsequent profit growth.



\subsection{Datasets}

\subsubsection{Polyvore Dataset}
The Polyvore dataset \cite{han2017learning} \cite{githubPD}was obtained from Polyvore.com, a well-known fashion website where fashion stylists can showcase their outfit creations to the public. The dataset has been previously employed in various studies related to fashion analysis.  

 There are datasets for training, validation, and testing the methods. Each of the datasets follow a similar hierarchical structure of organization -- each outfit contains multiple clothing items or accessories, and each individual item includes certain attributes, such as the name, price, likes, category id, and image of the item. 
 
 We were able to download these datasets directly without crawling, and there is a script \cite{githubPD} that forms the Fashion Graph using the raw JSON file. 

The Polyvore Dataset contains both image and textual description of the items. We use this to compare multimodal performance for one of our methods. Additionally, it contains curated outfits which can be used to train and validate the model. 
\subsubsection{Raw Data Statistics}

The Polyvore data consists of 21889 outfits over 380 categories, which were split into training, validation and test sets. To ensure the quality of the training, only the categories that appear in the sample more than 100 times were retained, leaving us with 120 categories. Furthermore, the outfits with less than 3 items were filtered to maintain uniformity. 

There are 126054 items overall, divided into 120 categories. Of these, 16983 outfits are from the training set, 1497 are from the validation set, and 2697 are from the test set. An outfit's average size is 6.2, and its largest size is 8.

Due to computational constraints with using COC ICE servers, we had to take a subset of the total data-set of about 15\% (train-val-test dataset). This leads to a sub-par training compared to the papers \cite{Cui_2019} \cite{fashion_hypergraph} we are replicating, this is also seen in section Results where the metrics reported don't match that of the original papers.

\section{Related Works}
The earliest works of outfit representation fall into two general categories. The first is a pair representation \cite{he2016learning,mcauley2015image,shih2018compatibility,veit2015learning}, in which the model generates compatible pairs of outfit items (e.g. a shirt and a necklace). Though this method was a great starting point, it failed to model the complex nature of fashion decisions, where often more than two pieces are involved.

The second, more advanced representation is sequential -- this allowed for three or more items to be paired together. Specifically, \cite{han2017learning} treated each outfit as a bidirectional sequence with a specific order (often from top to bottom, followed by accessories)), then they trained a Bidirectional Long Short Term Memories (Bi-LSTMs)  to progressively predict the next item conditioned on previous ones to learn their compatibility relationships given the clothing in an ensemble. An obvious limitation of this approach is that it does not use the structural information that graphical models use. However, it was restrictive because an outfit is a set of clothing, with no inherent sequential property.

This led to the innovation of utilizing graphs to represent individual items and outfits using GNNs\cite{gori2005new}. Each article of clothing is represented as a node in the graph, and an edge is an interaction between the different categories of clothing -- these components allow for outfits to be represented as subgraphs. After creating a fashion graph based on these rules, the authors were able to implement Graph Neural Network (GNN) techniques to create node embeddings and calculate compatibility between different clothing items. They were also able to test their model and create recommendations based on the GNN.

\section{Methodology/Experimental Setup}

\subsection{Evaluation metrics:}
We conducted experiments on two tasks: fill-in-the-blank fashion (FITB) recommendation and compatibility prediction. 

\subsubsection{Accuracy - Fill In The Blank (FITB) Task: }
This task helps in answering the question which item will complete an outfit. For example, given 3 pieces of items in an outfit, namely a shirt, hat, and shoes; we  test our model's performance in choosing an item which will complete our outfit from four options, assuming one option out of the four is more compatible (pants in this example). For each outfit, we randomly mask an item and test which item out of a subset will be most compatible with the existing outfit. We create a list of 3 negative items randomly sampled, and along with the masked item we have a 4 item set from which the most compatible item is chosen. We find outfit compatibility scores for all 4 sets of outfits built using the existing set of items in the outfit and the 4 choices. The choice with the highest outfit compatibility score is chosen as the answer. Since picking the right response from four pieces of fashionwear will be 25\% if chosen randomly, we can compare our model's accuracy against that.

\subsubsection{AUC - Compatibility Prediction: }
For the second task, the aim is to predict compatibility scores for any given outfit. Specifically, for each outfit in the test dataset, we create a negative outfit by randomly replacing one item with another random item from the corpus. After that, we use the popular and widely accepted AUC (area under the ROC curve) scores for evaluating our model. 

\subsection{Parameters:}
Due to space constraints, we worked with a subset of 1600 outfits which we split using a 70-30\% split into test and training data.
\\
We replicated the two methods using TensorFlow v1.15. We also used Google Inception and Hugging Face's Visual-Transformer (ViT) models for creating image embeddings.  All our experiments were run on the COC ICE server using RTX6000 GPUs. For optimization, we adopt the RMSProp and Adam optimizers. We early stop the training process when the loss stabilizes, usually around 15 epochs. 

\section{NGNN- Baseline Method}

\begin{figure}[h]
\includegraphics[width=\textwidth]{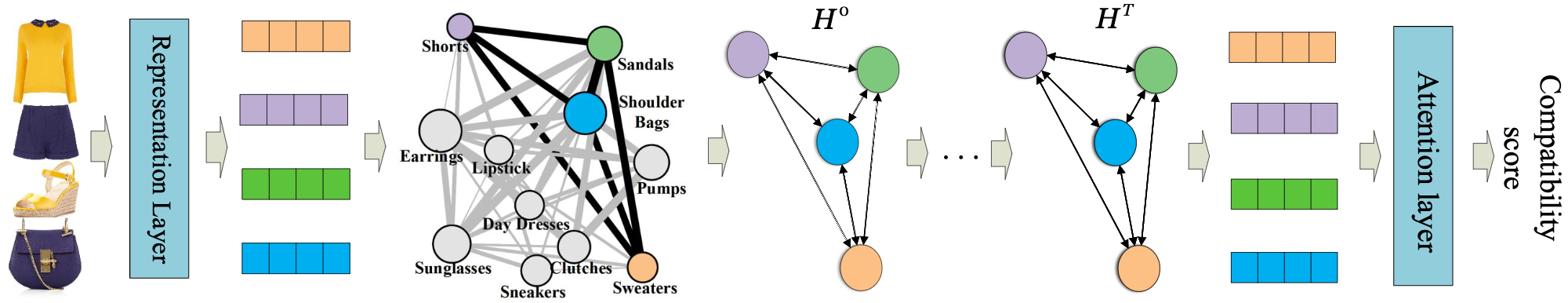}
 \label{fig:Framework}
\caption{NGNN Framework}
\end{figure}.
\subsection{Introduction}

The Node-wise Graph Neural Networks\cite{Cui_2019} approach was the first paper to utilise graphs for solving the outfit compatibility problem. By representing categories as nodes in a fashion graph, researchers were able to tackle many challenges that only graphs could address. First, they were able to circumvent the problem of variable length input by representing outfits as a subgraph. Though this problem was addressed using Bi-LSTMs \cite{han2017learning}, which represent an outfit as a sequence, Bi-LSTMs introduced a bias because outfits do not inherently possess a sequential structure. Additionally, by representing an outfit as a subgraph, the NGNN approach allowed for the modelling of more complex interactions between clothing/items and outfits. Finally, it made the category of an item a mainstay of the model in a number of ways, like a) using category-dependent weighted message passing aggregation b) using category-specific feature mapping for item embeddings.

The code repository for the paper was available to us via \href{https://github.com/CRIPAC-DIG/NGNN}{GitHub}. However, the code needed to be adapted to the latest Tensorflow version as it was written in Tensorflow 1.15. We also implemented code to extract embeddings of the images of items using a pre-trained Vision Transformer to compare results.
\subsection{Technical description}
The paper uses the 120 clothing/accessory cate  gories as nodes with an edge existing between them if they co-occur in an outfit. The training strategy discussed below helps to bring the process of representing an outfit as a subgraph to life. It also illustrates how this subgraph can be distilled to a single score using an attention mechanism.

\begin{figure}[h]
\includegraphics[width=\textwidth]{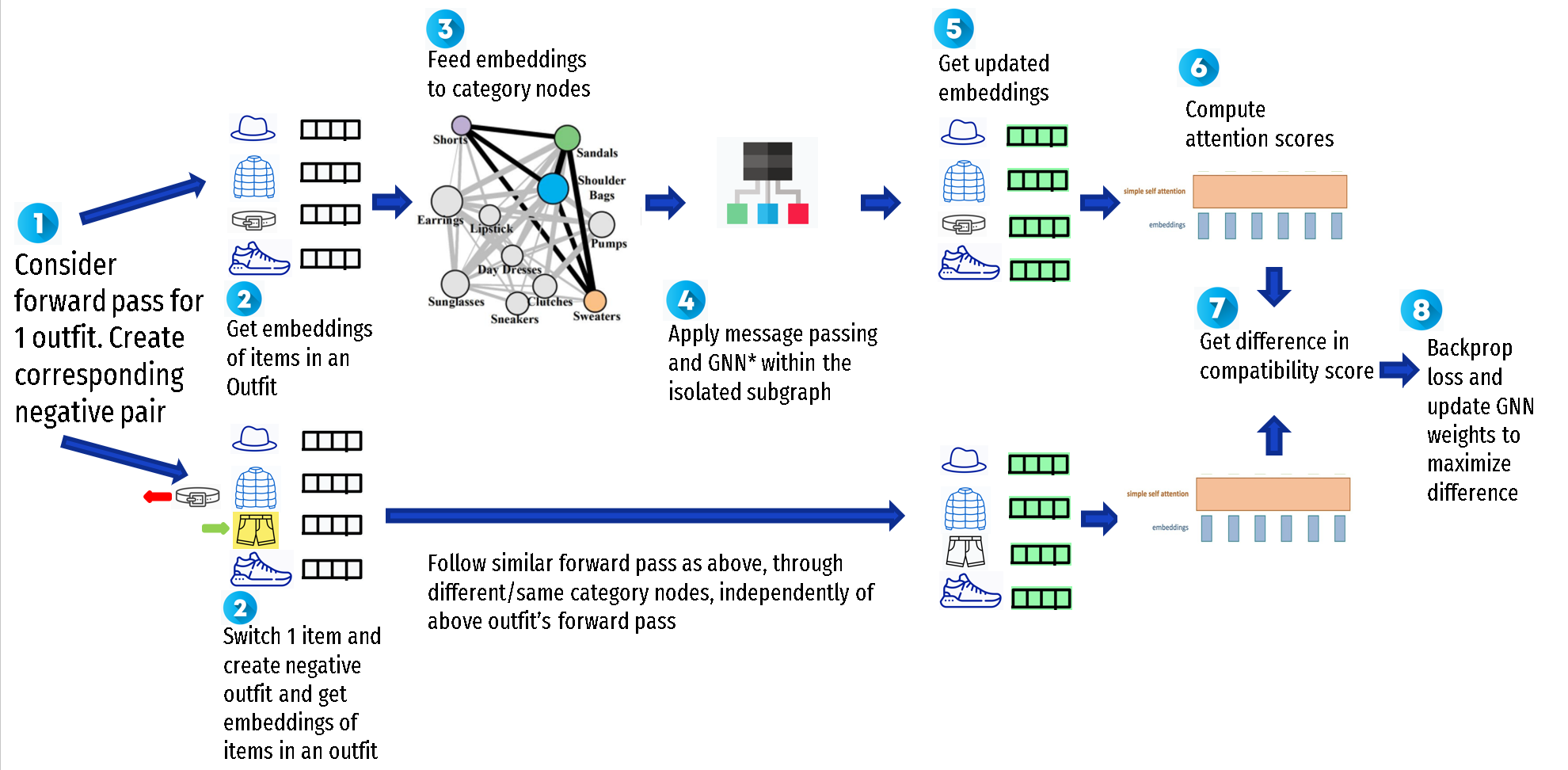}
 \label{fig:fig2}
\caption{NGNN Training Flow}
\end{figure}

The following describes one iteration of a forward pass by passing a single outfit as input. As a reminder, an outfit consists of a list of 7-8 items for each of which we have a text description and an image.

The embeddings of the text description are obtained using the one-hot encoding(2757-dimensional Boolean vector) after the pre-processing text. The embeddings of the image are created using a pre-trained InceptionV3 model(2048-
dimensional vector).

\begin{itemize}
  \item Step 1: For an outfit, we create a corresponding "negative" outfit. This is done by randomly selecting an existing item in the outfit and replacing it with a randomly sampled item. Then, the next steps are performed for both the positive and negative outfits.
  \item Step 2:  We then fetch the embeddings of each of the items(text/visual depends on the mode, for now, assume visual embedding only). 
  \item Step 3: Pick the category nodes that the items belong to and initialize the feature vector of these nodes with the embeddings of the items to each of the respective category nodes. 
  \item Step 4: Then, we start the message passing and update the embeddings of each of the items. 
  \item Step 5: Next, we obtain the updated item embeddings. 
  \item Step 6: Finally we use these embeddings of an item and feed it to an attention layer to compute attention scores. These are the outfit compatibility scores for this outfit.
  \item Step 7: Now we obtain the difference of 2 scores of the positive and negative outfit 
  \item Step 8: Finally we get the loss and backpropagate through the network to update the attention layer and GNN weights.  
\end{itemize}
It's important to note that the rest of the category nodes of the graph are completely ignored and all further message passing is done only between the existing nodes.

The hyper-parameters were determined by the grid search strategy. These include learning rate, batch size, visual-text trade-off parameter $\beta$, the parameter for L2 regularization $\lambda$, hidden size d and the number of message propagation steps T.  The model achieved optimal performance
with the learning rate as 0.001, batch size as 16, $\beta$ as 0.2, $\lambda$ as 0.001, d as 12 and T as 3. After subsetting the input dataset, we were able to run around 12 epochs for the training process for HGNN and around 4 epochs for NGNN. NGNN was significantly slower to train than HGNN even after tweaking the learning rate. All the experiments were conducted using GPU RTX-6000.


\section{Hypergraph Neural Network}

\subsection{Introduction}

Another method that was explored to solve the problem of outfit compatibility involves leveraging hypergraphs \cite{fashion_hypergraph} to represent outfits and outfit categories. A hypergraph is similar to any other graph, but has a key difference that an edge can connect multiple nodes -- these special edges are called "hyperedges." The code repository for the paper was available to us via \href{https://github.com/outfit-net/outfit-commpatibility/tree/main/OCPHN}{GitHub}.


A hypergraph can encode high-order data correlation (beyond paired connections) utilizing its degree-free hyperedges, in contrast to a simple graph, in which all edges must have a degree of 2, as shown in Figure 2. The adjacency matrix in Figure 2 is used to represent a generic graph, where each edge only connects two vertices. A hypergraph, on the other hand, can be easily enlarged for multi-modal and heterogeneous data.

\begin{figure}[h]
\includegraphics[width = \textwidth]{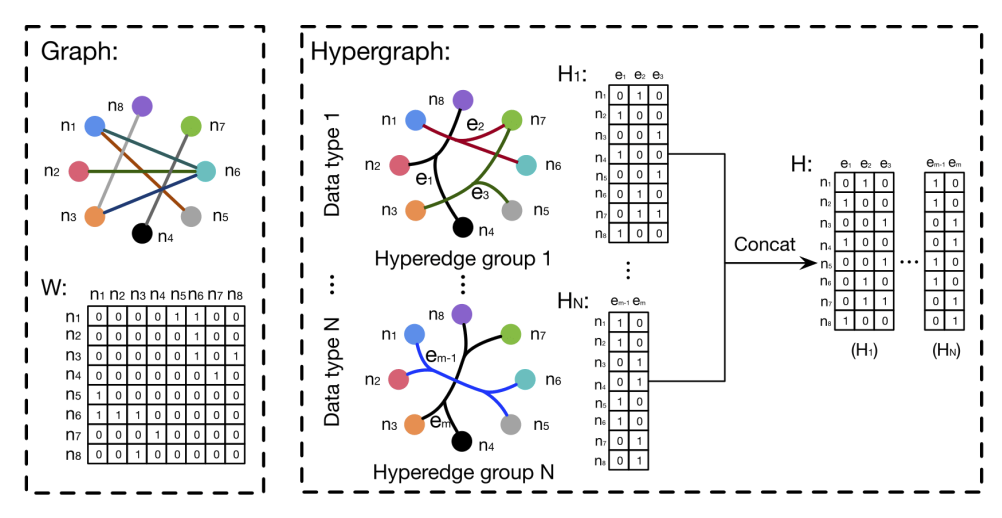}
 \label{fig:Framework}
\caption{Comparision between HGNN and GNN}
\end{figure}.

\subsection{Hypergraph Creation}

As shown in Figure 2, we created a fashion hypergraph H = (V, E) based on the training dataset where the category information of the items acts as the prior knowledge of the items. The hyperedge (i.e., the linkages of the same colour) represents the interaction between several categories when they occur in the same outfit, whereas each node is specifically linked to only one unique category. Therefore, each outfit may be thought of as a hyperedge of a hypergraph if each component is filled into its respective nodes.

In this framework, each clothing/accessory item category (e.g. black pants) is represented as a node, and several nodes are connected by a hyperedge if the categories represented by the nodes constitute an outfit.

\begin{figure}[h]
\includegraphics[width = \textwidth]{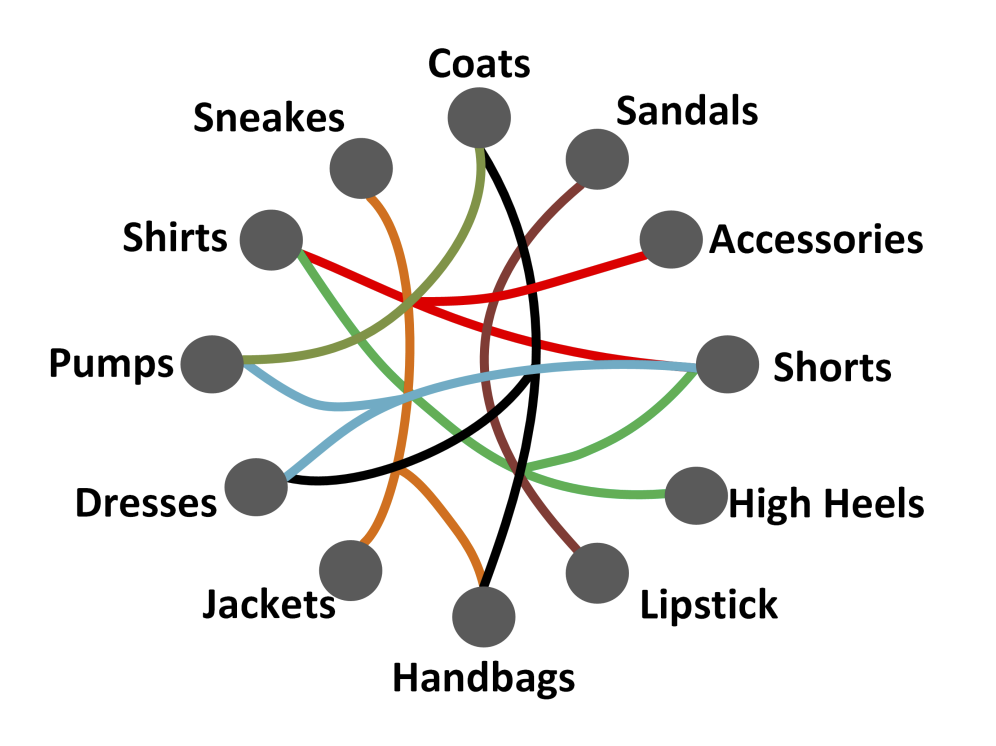}
 \label{fig:fig2}
\caption{Sample Hypergraph in the Polyvore data }
\end{figure}

\subsection{Framework}

\begin{figure}[h]
\includegraphics[width =  \textwidth]{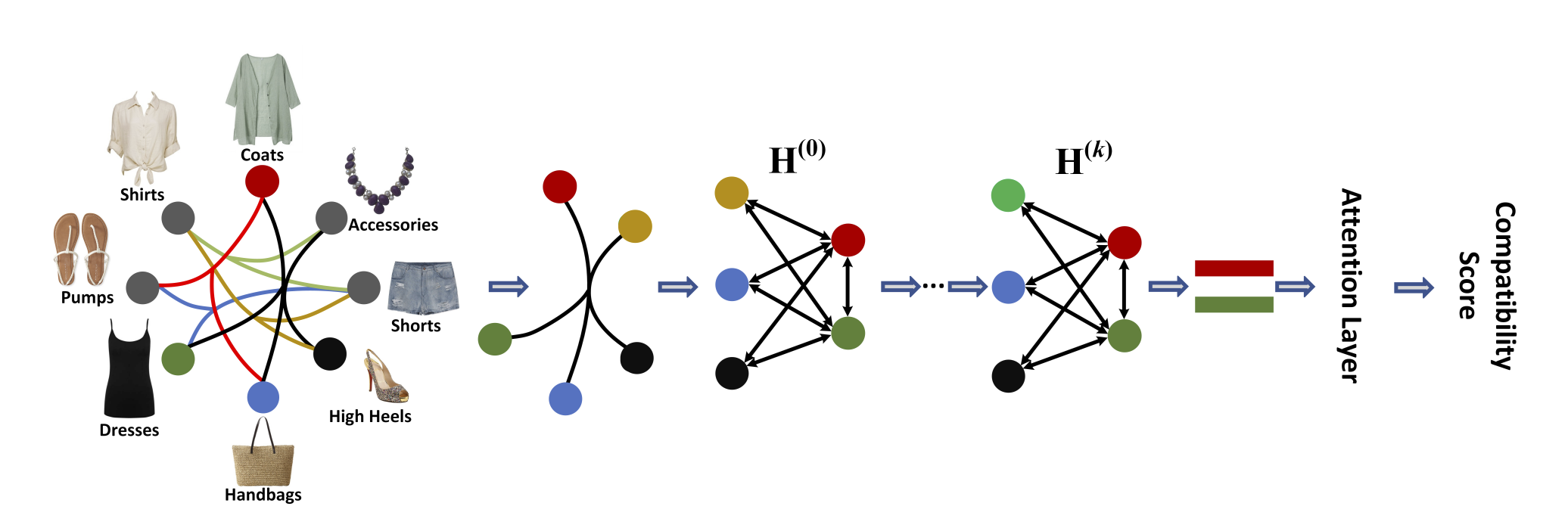}
 \label{fig:fig2}
\caption{Hypergraph Framework}
\end{figure}.

Only two nodes are chosen for each hyperedge during the transformation process to represent the full hyperedge. The remaining nodes, referred to as mediator nodes, are linked with the key nodes to create a new graph in order to fully utilize all of the nodes' information. Each mediator node connects two essential nodes, as seen in Figure 4.

This is similar to the previously discussed NGNN approach in that it utilizes item categories as nodes, but instead of outfits being represented as sub-graphs, outfits are simply connected by hyperedges.  

As outlined in the steps for the NGNN framework, the node embeddings in step 3 are generated from the hypergraph structure.

The hyper-parameters were determined by the grid search strategy. These include learning rate, batch size, visual-text trade-off parameter $\beta$, the parameter for L2 regularization $\lambda$, hidden size d and the number of message propagation steps T.  The model achieved optimal performance
with the learning rate as 0.001, batch size as 16, $\beta$ as 0.2, $\lambda$ as 0.001, d as 12 and T as 3. After subsetting the input dataset, we were able to run around 12 epochs. NGNN was significantly slower to train than HGNN even after tweaking the learning rate. All the experiments were conducted using GPU RTX-6000.

\section{Experiments and Results}
\subsection{Comparing Model's Performance}
The results of our experiments are summarized in the table below:
\begin{table}[!ht]
    \centering
\begin{tabular}{ccc}
\hline Method & Accuracy (FITB) & AUC (Compatibility) \\
\hline Random & $24 \%$ & 0.51 \\
NGNN & $38 \%$ & 0.65 \\
HGNN & $39 \%$ & 0.76 \\

\hline 
NGNN (with VIT) & $40 \%$ & 0.68 \\
HGNN (with VIT) & $40 \%$ & 0.77 \\
\hline
\end{tabular}
    \caption{Performance Comparison} \label{table1}
\end{table}
\\
As mentioned in Dataset Section, we use a subset of the dataset (both for training and validation and testing). This leads to our accuracy scores being lower than what was reported in the original papers \cite{Cui_2019}
\\

\subsubsection{FITB Accuracy:}

The comparison of FITB accuracy is shown in the middle column of Table. We can see that the random selection model gives FITB of 25\% as expected ( as there is a 1 in 4 chance of picking the right item randomly). Both NGNN and HGNN perform almost similarly with the FITB task, with HGNN being slightly better presumably because hypergraphs can capture more complex interactions between nodes.  \\
We also tried building better and more space-efficient embeddings using the latest Visual transformer model and using that on our two models gave only a slight improvement in the accuracy on test data. 

\subsubsection{AUC Outfit prediction accuracy:}

This is shown on the last column of Table. The comparison across different models is similar to the FITB accuracy metrics. HGNN achie

\subsection{Testing impact of Modality:}
As we have both textual descriptions of our items as well as images, we can compare the performance of models built using different modalities.
This is summarized in the table below: 

\begin{table}[!ht]
    \centering
\begin{tabular}{ccc}
\hline Method & Accuracy (FITB) & AUC (Compatibility) \\
\hline 
NGNN (visual) & $35 \%$ & 0.62 \\
NGNN (textual & $37 \%$ & 0.63 \\

\hline 
NGNN (multi-modal) & $38 \%$ & 0.65 \\

\hline
\end{tabular}
    \caption{Impact of different modalities} \label{table2}
\end{table}
The text-Only model uses features created using detailed textual descriptions of each item. Embeddings created using that were fed into the NGNN model.
Similarly, the visual-only model uses embeddings from the images of items only. The muli-modal model uses both of those embeddings to train the NGNN model.
\\
As expected the multi-modal performs best (as seen in Table 2). Though the text-only model performs slightly better than the visual-only model. This could be because the detailed descriptions of each item in an outfit and the embeddings created from them are concise enough to capture all key features of the item.

\section{Conclusion}

During our experiments, we observed that GPT-annotated fine-tuning produces equivalent results to human annotated fine-tuning in the multi-class classification task( 79.5\% for GPT-annotated as compared to 82.8\% for human annotated), whereas the performance deteriorates drastically in the question answering task (33.6\% accuracy for GPT-annotated as compared to 48.99\% for human-annotated). We believe that the difference in performance is because of the difference in the objective tasks. The question-answering task is complex as compared to the classification task and requires a deeper understanding of the training data during fine-tuning. Thus, it is important that training data during fine-tuning is as accurate as possible. Further, the overall performance of the question-answering task is below par with the performance of the classification task. This might be because of limited data and a smaller number of epochs in training.

Future work can include experimenting with larger data, training for higher number of epochs, experimenting with prompts for generating more accurate synthetic response, and identifying better evaluation metric for subjective question-answering.

\bibliographystyle{iclr2021_conference}
\bibliography{iclr2021_conference}


\end{document}